\documentclass[conference]{IEEEtran}

\usepackage{graphicx}
\usepackage[colorinlistoftodos]{todonotes}

\title{Game AI Research with Fast Planet Wars Variants}

\author{Simon M. Lucas \\ Game AI Research Group \\ 
School of Electronic Engineering and Computer Science \\
Queen Mary University of London}

\begin{document}

\maketitle
\thispagestyle{empty}
\pagestyle{empty}

\begin{abstract}

This paper describes a new implementation of Planet Wars, designed 
from the outset for Game AI research.  The skill-depth
of the game makes it a challenge for game-playing agents, and
the speed of more than 1 million game ticks per second enables
rapid experimentation and prototyping.
The parameterised nature of the game together with an interchangeable
actuator model make it well suited to automated game tuning.
The game is designed to be fun to play for humans, and is directly playable
by General Video Game AI agents.

\end{abstract}

\section{Introduction}

This short paper describes a new platform for Game AI research based on variations of the Planet Wars game.  The platform has been designed from the ground
up for speed and flexibility.  The core game has a fast forward model and an efficiently copyable game state,\footnote{Some game 
implementations have state variables spread throughout the code making
it awkward and possibly inefficient to copy the state.} and is therefore
ideal for statistical forward planning algorithms.
Furthermore, it has many options which are all bundled
into a single parameter object, so that the details
of the game can be changed dynamically
within different copies of the game.

This enables the effects of inaccurate forward models
to be systematically investigated, and also makes it
well suited to research into automated game tuning.  The 
games are directly playable by General Video Game AI (GVGAI)
agents, therefore adding games with strategic depth
to that platform.  This is particularly useful
for extending the type of games offered by the 
GVGAI 2-player track \cite{GVGAI2Player},
but can also be used for the single player tracks \cite{GVGAISurvey}
by providing one or more fixed opponents.
Finally, the game has also been
designed to be fun for human players.

The last decade has seen increasing interest
in testing AI on Real-Time Strategy games, with
StarCraft being an obvious example.  There is also
an important place for simpler and faster games,
which not only offer more convenient experimentation
but can be easily varied to provide more general and
varied AI challenges than can be offered by a single game.
Recent examples include microRTS \cite{ontanon2017combinatorial} and ELF \cite{DBLP:journals/corr/TianGSWZ17}.
Compared to microRTS and ELF, the game described
in this paper runs around ten times faster due to the 
simpler rules of the game, and specifically due
to design decisions which limit the number of game entities
in play at any one time.  
Compared to the stripped down
version described in \cite{lucas2018n} the platform in this paper offers
more sophisticated game play as described below,
and introduces the additional features of spinning turrets
and a gravity field.

\section{Planet Wars}

Planet Wars is a popular casual two-player real-time strategy game with versions going under many names on various platforms.
The game is a simple real-time strategy game that is fun for humans to play and provides
an interesting challenge for AI.  
It was used for the 2010 Google AI challenge  (http://planetwars.aichallenge.org/)
run by the University of Waterloo in Canada ~\cite{fernandez2011optimizing} with great success.
 The game was also successfully used by Buro et al.\
for a Dagstuhl AI Hackathon, who also describe
the rules of the standard game  \cite{lucas2015Dagstuhl}.

\subsection{Rules}

The aim of the game is for a player to take over all enemy planets by sending ships to invade them.  Each planet
is either owned by either player or is neutral, and
each non-neutral planet spawns new ships.  Good
strategy involves balancing the need to take over 
as many planets as quickly as possible versus leaving
currently owned planets with enough ships to deter 
invasion.  The game is played out on a 2D map, and 
ships take time to travel between planets, hence there
is an interesting spatio-temporal planning aspect to the
game.  Indeed, Leece and Jhala \cite{JhalaPlanetWars} 
specifically chose the game to study the ability of q-learning agents to deal with spatial planning.

The Map specifies how the planets are laid out in 2D space.
Planets are defined by their position, their size,  growth rate (proportional to their radius), 
and initial ownership.
Hence the map alone already provides for immense variations: 
it is easy to generate new maps at random that will
require different tactics in order to win.  All aspects of the map play an important role in the decision making process.  Hence, even in its default setting the game
already offers significant variation, and offers
a more robust test of an AI system than the
individual games of the Atari 2600 platform \cite{MachadoALE2017}, for example, even before we 
consider the actions of a varied set of adversaries.  

Beyond this, the time taken for ships to travel between planets 
also affects game play, making ship speed and map size important
parameters.

Further variations are possible with the user-interface.  
In some versions the player selects a single source and target (destination) 
planet for each move, in other versions multiple source planets can be selected 
via a drag action.  However, most versions the author is aware of use the 
basic source / destination mechanism as a way to specify actions.

\subsection{Variations}

All the above listed features are standard for the game.  For this
version we introduce a number of variations.  The first two
are for the purpose of efficiency.  Firstly, a transporter is used
to send ships between planets, so multiple ships travel on a single 
transporter.  This is more efficient than sending them individually.  Secondly,
each planet is restricted to having a single transporter.  This means the
cost of calculating game updates via the next state function grows only
linearly with the number of planets in the game.  For comparison,
the version used for the Google AI challenge bundled multiple ships
on to a single transporter, but placed no limit on the number of
transporters that could be launched at any time, other than the constraint
that each one must carry at least one ship.

Beyond efficiency, two significant variations are introduced to
add to the game-play: the gravity field
and the rotating turret for direction selection.

The gravity field pre-computes a gravitational force that is calculated from the 
position and mass of each planet with mass being proportional to the 
planet's area (since this is 2D space).	 The gravity field adds significantly to
the game-play: the fact that
ships now follow curved trajectories means that players (especially 
when playing with the Slingshot actuator, see below) 
must carefully judge the effects of gravity 
when timing the release of the transporter.
The curved trajectories may be more interesting to observe than linear
ones, and create some dramatic tension around whether the transporter will
reach the targeted planet or not.

The rotating turret adds a skill aspect for human players.  Using this mechanism a player executes a long mouse press
on a planet for each action.  
While the mouse is pressed, ships are loaded on to a transporter at each game tick.  When the mouse is released, the transporter leaves heading in the current direction of the turret.

For a human player this added skill can be a source of enjoyment or of frustration, depending on the individual player and on how well the game is tuned for that player.  For example, if the turret rotates too quickly then it becomes difficult to target the desired planet; if too slowly, then a long wait may be involved for the turret to point in the desired direction.  For an AI agent, a slow turret can be a source of challenge since it requires planning further ahead, whereas a fast rotating turret should not make it more difficult, unless timing noise is added to the AI agent's actions.


\subsection{Game Parameters}

The game currently has 16 parameters which significantly affect the game play.  At the time of writing these include the following:

\begin{itemize}
\item Number of planets: more planets lead to higher branching factors and more complex game-play.
\item Map dimensions: the width and height of the map in pixels.
\item Gravitational constant: multiplies planet mass to scale gravitational force.  Higher values lead to more curved transit trajectories.
\item Growth rate range: planet growth rates are sampled from a uniform distribution in this range.
\item Radial Separation: as planets are placed randomly they must be at least this number of radii apart from the nearest already placed planet.
\item Ship launch speed: faster launch speed means ships will tend to arrive at their destination sooner, and less influenced by the gravity field.
\item Transport tax:  a subtractive amount per tick that reduces the number of ships during transit (and may even take them negative, hence turn them in to opponent ships.
\end{itemize}

\begin{figure}[hbtp]
\centering
\includegraphics[width=0.9\linewidth]{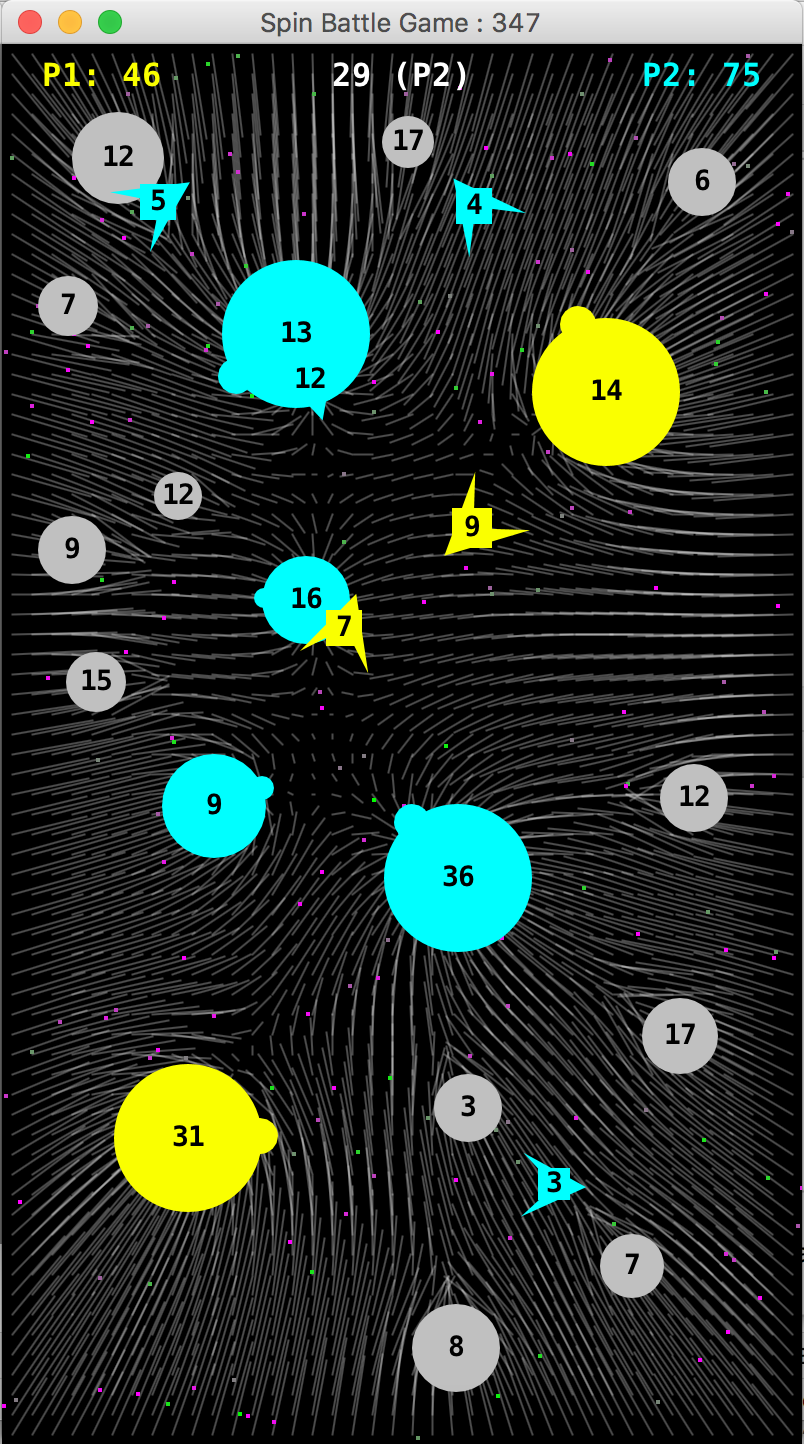}
\caption{\label{fig:SpinBattlePortrait}
Planet Wars with spinning planets and a gravity field,
which both have significant effects on the game play.}
\end{figure}

A screenshot of the game is shown in figure~\ref{fig:SpinBattlePortrait}.
This version is shown in portrait mode; the dimensions
of the game can be changed as easily as any other parameter, though there
are dependencies on other game parameters: for example, making the
screen area smaller while retaining the same planet size will affect
how many planets can be placed, and the density of the area.  Maps
with denser layouts may place greater importance on owning well-connected
central planets.  The ability to easily change screen dimensions may
be true of most games with randomly generated levels, but does not
hold for games that rely on file descriptions of each level, 
such as Pac-Man and Super Mario Bros.\ 
(for those games, level generation is a topic in its own right).

\subsection{AI Agent Considerations}

AI agents can submit at most one action per game tick, though the details depend
on the actuators used, described in section~\ref{s:implementation}.
Currently there is no fog of war: all game states are fully observable 
to the AI agents, though this is an obvious source of future variations.
An existing variation shows players only the ownerships of each neural or
opponent planet, and not the number of ships on them.  Other observability
variations are also possible, such as just showing a small window of the 
map instead of the entire map.  For an example of how to explicitly vary
observability in video games see the Partially Observable Pac-Man competition
\cite{PO-Pacman}.

The branching factor of the game (average number of legal actions at each tick) 
depends very much on the actuator model: for the source / target model,
the source planet must be owned by the player, but the destination can
be any planet other than the source.  The number of the planets is a parameter
of the game, and during testing this has been varied between 10 and 100.
Games may last for many thousands of ticks, but for experiments we
often limit this to between 1,000 and 5,000 ticks, which is often enough
to estimate which player is superior.  Games between a strong and a weak player
are often decided (and terminated) within 1,000 ticks.

We do not have statistics to support this yet, but the game 
seems to proceed in distinct phases.  In the initial phase each player
owns a small number of planets and the game appears to be finely balanced: 
decision on which ones to invade at this stage are
important, though we do sometimes observe AI players throwing away an
apparent lead.  In the next phase the lead frequently fluctuates with many ships
in transit and closely fought battles.
This is followed by a final phase where one player dominates and the outcome of the
game is no longer in doubt.

The game was recently tested during a graduate level AI Assisted Game Design
course.\footnote{https://github.com/GAIGResearch/AIGD2}  
During the course students developed their own statistical forward planning agents
and also their own tuned versions of the game, by adjusting parameters 
and varying the rules.  All game variants were able to clearly separate the
agents in to different levels of skill, measured by win rates in round robin
leagues (leagues in which each player plays every other player a fixed number of times), providing evidence that the game
variants have skill depth.  Interestingly, the game variants interesting produced very different ranking of the AI agents.

\section{Implementation}
\label{s:implementation}

The game is implemented in Java, and has been designed from the
ground up to run efficiently, offer flexibility for the Game AI researcher, 
and also allow for easy human interaction.  Enabling easy human interaction 
enables automatically tuned games to be tested by human players.
Efficiency is important for all aspects of the research.

Key features of the design include:

\begin{itemize}
\item All variable parameters are stored in a single GameParameter object, and are never declared as static variables.  A reference to a GameParameter object is passed to all copies of the game, but can be copied and modified on demand.
\item Each planet has only a single Transporter (these are the angular space ships shown journeying
through space in figure~\ref{fig:SpinBattlePortrait}).  This enables the cost of game state copying and updating to be kept linear rather quadratic in the number of planets.  This does not seem to have a detrimental effect on the game play.
\item The game state contains no circular references and can therefore be serialised into JSON for convenient storage and transmission.  The largest part of the game state is the Gravity Field (which is a 2D array of Vector2D objects), but this can be nullified for serialisation and re-created on demand when needed.
\item The \emph{actuator} model has been decoupled from the game state.  This means that different ways of controlling game actions can be plugged in.
\end{itemize}

Regarding the last of these points, so far two actuators have been implemented.  
An additional actuator based on a directional catapult is planned.

\subsection{Source Target Actuator}

Each move consists of a source planet being selected, followed by a destination planet.  The move is executed only if the source planet's transporter is currently at home, and the planet is owned by the player.  If these conditions are satisfied, then the ship is loaded with a percentage of the planet's ships, and launched in the direction of the target planet.

AI players currently see the number of actions at each tick as being equal to the number
of planets.  Low-level actions are grouped in to source-target pairs,
with illegal actions (ones in which the player does not own the source planet) 
being ignored.  A video of a rolling horizon evolution agent playing against a hand-coded heuristic agent can be viewed here: 
{\bf https://www.youtube.com/watch?v=G2aoxYODs9U.}

\subsection{Slingshot Actuator}
Each move consists of a user selecting a planet for a number of game ticks.
The selection only happens if the player owns the chosen planet.
When the planet is deselected, the ship is launched at a standard speed
in the direction the turret is facing.
A video of a human player (the author) playing against a heuristic agent player can be viewed here: {\bf https://www.youtube.com/watch?v=y2q5VW8kS8k.}

\subsection{Timing Results}

\begin{table} [!t]
\centering
\caption{\label{tab:speed}Speed of key operations in units of thousands of operations per second (iMac with 3.4 GHz Intel Core i5 CPU).  Note that the Gravity Field (GF) is just computed once at the start of each game, after the position and size of each planet has been fixed.}
\begin{tabular}{lrr}
\hline
Operation & kop/s (1) & kop/s (4) \\
\hline
nextState & 870 & 1,640 \\
copy & 1,600 & 3,230 \\
compute GF &  1 & 2 \\ 
\hline
\end{tabular}
\end{table}

Table~\ref{tab:speed} shows the timing results for a single thread and four threads running on an iMac with core m5 processor.  The software includes the facility to run multiple games in multiple threads, or different game agents in different threads, enabling speeds in excess of 1.6 million ticks per second when running four threads simultaneously.  For comparison, ELF and microRTS offer speeds of around $50k$ ticks
per second when running single-threaded, meaning that this game is more than 10 times faster.  The games are obviously different so comparing timings
may seem unfair, but the point of the comparison is to highlight the
speed offered by our platform and the rapid generation of results
that this enables, even on a standard laptop or desktop computer.

\section{Generating AI Agent Results}


The software distribution includes the following experiments ready to run, including
human versus AI and AI versus AI.  For human versus AI, the AI controllers are generally
superior to casual human players, but their intelligence can easily be varied
and decreased if necessary by reducing the simulation budget or the sequence or rollout length,
in order to provide an easier challenge.

For AI versus AI, the game has been tested on the graduate student course mentioned above,
and also for this paper a small but representative test was run using 3 different controllers, reported in
table~\ref{tab:agents}.
RHEA is the rolling horizon agent from Lucas et al \cite{lucas2018n}, but with a sequence length of 200 and 20 iterations per move. The MCTS agent is the sample agent from the GVGAI 2-player track, but with a rollout length set to 100 and iterations per move set to 40. 
Hence both RHEA and MCTS used an evaluation budget of approximately $4,000$ game ticks per move.  Rand is a uniform random agent.  Games were limited to $2,000$ moves but often terminated with a win before reaching the limit. Running the 60 games for this mini-tournament took less than 5 minutes on the iMac computer described above.  
A follow-up paper will investigate these results more thoroughly, 
but recent experiments on this and on some other games show rolling horizon evolution
frequently outperforming MCTS.

\begin{table} [!t]
\centering
\caption{\label{tab:agents} Results of playing three agents
against each other in a round-robin league on 10 fixed maps,
playing each map twice, so 60 games in total.  See text for description of each agent.  The table shows the row agent wins against the column agent, with the rightmost column showing the total number of wins for the row agent.}
\begin{tabular}{lrrrr}
\hline
 & RHEA  & MCTS & Rand & Wins \\
\hline
RHEA & - &  18 & 20 & 38 \\
MCTS & 2 & - & 17  & 19 \\
Rand & 0 & 3 & - & 3  \\ 
\hline
\end{tabular}
\end{table}



\section{Conclusions}

The Planet Wars platform described in this paper provides
a useful addition to a growing number of games designed for AI research.  
The platform is efficient and well suited
to testing statistical forward planning algorithms such as Monte Carlo Tree Search and Rolling Horizon Evolution.  
The game already has sixteen parameters that can be varied in order to significantly affect the game play 
and provide a thorough test of the strengths and weaknesses of the competing agents.  


Future work includes introducing further variations while
retaining the speed of the game, integration with
AI environments such as OpenAI Gym, and additional actuator models such as a directional catapult.  
The speed of the game and its extensive parameter set also make it well suited to automated game tuning \cite{PowleySemiAutoGameTune,MikeGameTuneNTuple}.

\addtolength{\textheight}{-12cm}   





\bibliographystyle{IEEEtran}

\bibliography{spindemo}

\end{document}